\documentclass{bmvc2k}

\usepackage{times}
\usepackage{epsfig}
\usepackage{graphicx}
\usepackage{amsmath}
\usepackage{amssymb}
\usepackage{dirtytalk}
\usepackage[dvipsnames]{xcolor}
\usepackage{graphicx}
\usepackage[subtle]{savetrees}
\usepackage{color, colortbl}
\usepackage{soul}

\newcommand{\bv}{\mathbf{v}}

\newcommand{\bt}{\mathbf{t}}

\newcommand{\prepad}{\hspace{1.5em}}
\newcommand{\prepadmini}{\hspace{1em}}

\makeatletter
\newcommand{\printfnsymbol}[1]{%
  \textsuperscript{\@fnsymbol{#1}}%
}
\makeatother

\newcommand\blfootnote[1]{%
  \begingroup
  \renewcommand\thefootnote{}\footnote{#1}%
  \addtocounter{footnote}{-1}%
  \endgroup
}

\title{Use What You Have: Video Retrieval Using Representations From Collaborative Experts}
\addauthor{Yang Liu*}{yangl@robots.ox.ac.uk}{1}
\addauthor{Samuel Albanie*}{albanie@robots.ox.ac.uk}{1}
\addauthor{Arsha Nagrani*}{arsha@robots.ox.ac.uk}{1}
\addauthor{Andrew Zisserman}{az@robots.ox.ac.uk}{1}

\addinstitution{
 Visual Geometry Group\\
 University of Oxford\\
 UK
}

\runninghead{Liu, Albanie, Nagrani, Zisserman}{Collaborative Experts}

\begin{document}

\maketitle
\blfootnote{\hspace{-0.55cm} Details of the results correction can found in the appendix.}
\blfootnote{\hspace{-0.55cm} *Equal contribution.}

\begin{abstract}

 The rapid growth of video on the internet has made searching
 for video content using natural language queries a significant
 challenge. Human-generated queries for video datasets `in the wild'
 vary a lot in terms of degree of specificity, with some queries
 describing `specific details' such as the names of famous identities,
 content from speech, or text available on the screen. Our goal is to
 condense the multi-modal, extremely high dimensional information from videos into a single, compact video representation for the task of video retrieval
 using free-form text queries, where the degree of specificity is
 open-ended. 

For this we exploit existing knowledge in the form of pre-trained semantic embeddings which include `general' features such as motion, appearance, and scene features from visual content. We also explore the use of more `specific` cues from ASR and OCR which are intermittently available for videos and find that these signals remain challenging to use effectively for retrieval. We propose a \textit{collaborative experts} model to aggregate information from these different pre-trained experts and assess our approach empirically on five retrieval benchmarks: MSR-VTT, LSMDC, MSVD, DiDeMo, and ActivityNet.  Code and data can be found at \url{www.robots.ox.ac.uk/~vgg/research/collaborative-experts/}. This paper contains a correction to results reported in the previous version.
 \end{abstract}
\section{Introduction} 
 
Videos capture the world in two important ways
beyond a simple  image: first, video contains temporal information --
semantic concepts, actions and interactions  evolve over time; Second,
video may also contain information from multiple modalities, such as
an accompanying audio track. This makes videos both richer and more informative, but also more
challenging to represent.
Our goal in this paper is to embed the information from multiple modalities 
and multiple time steps of a video segment into a compact fixed-length
representation. Such a compact representation can then be used for a
number of video understanding tasks, such as video retrieval, clustering and summarisation. In particular, we focus on retrieval; our objective is to be able to retrieve
video clips using a free form text query that may contain both general and specific information.

\begin{figure}
    \centering
    \begin{minipage}{0.38\linewidth}
        \vspace{-1.2cm}
        \hspace{0.2cm}
        \begin{subfigure}{}
        \includegraphics[width=0.98\textwidth,trim={0 0cm 0 0cm},clip]{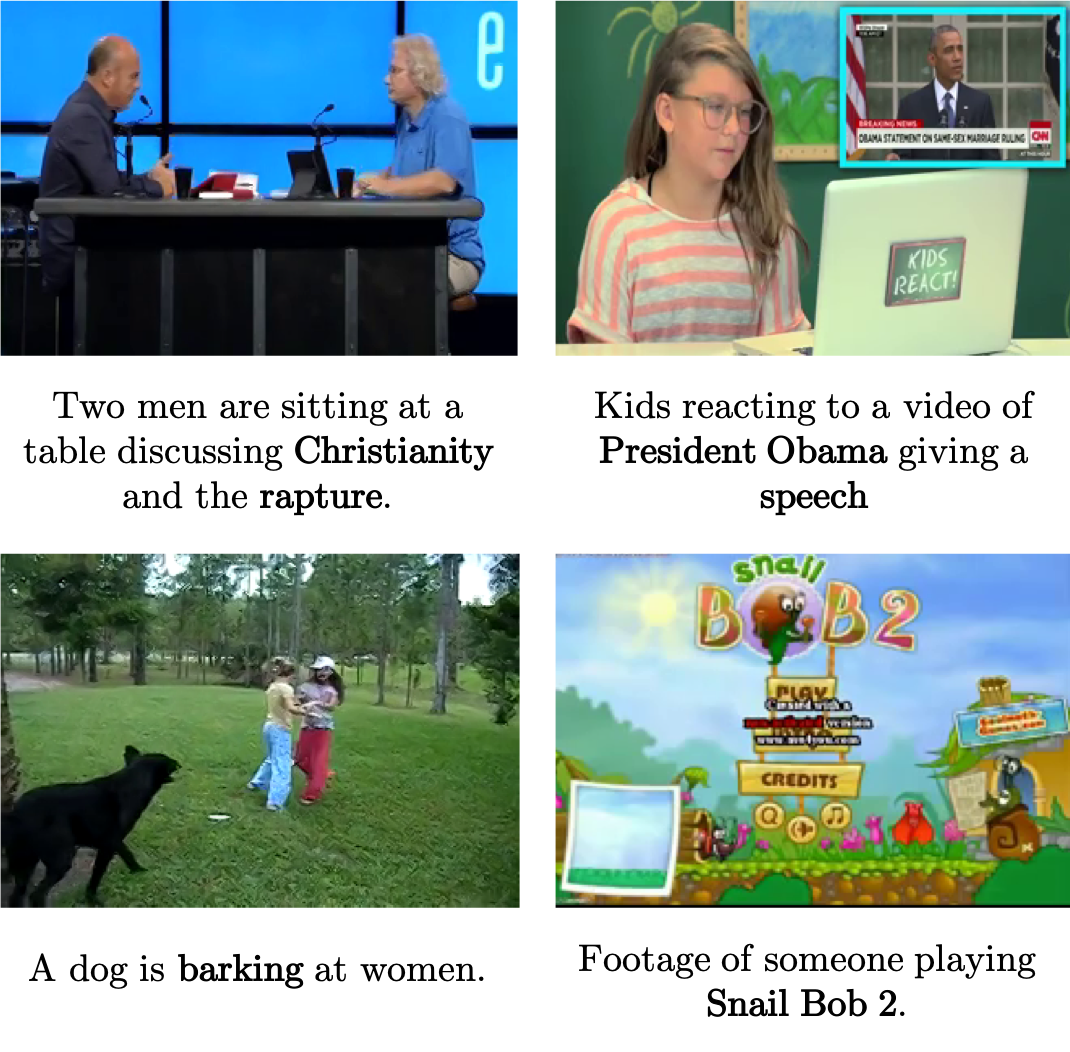}
        \end{subfigure}
    \end{minipage}%
    \begin{minipage}{0.6\linewidth}
        \hspace{0.2cm}
        \begin{subfigure}{}
            \includegraphics[width=\textwidth,trim={0 0.1cm 0 2cm},clip]{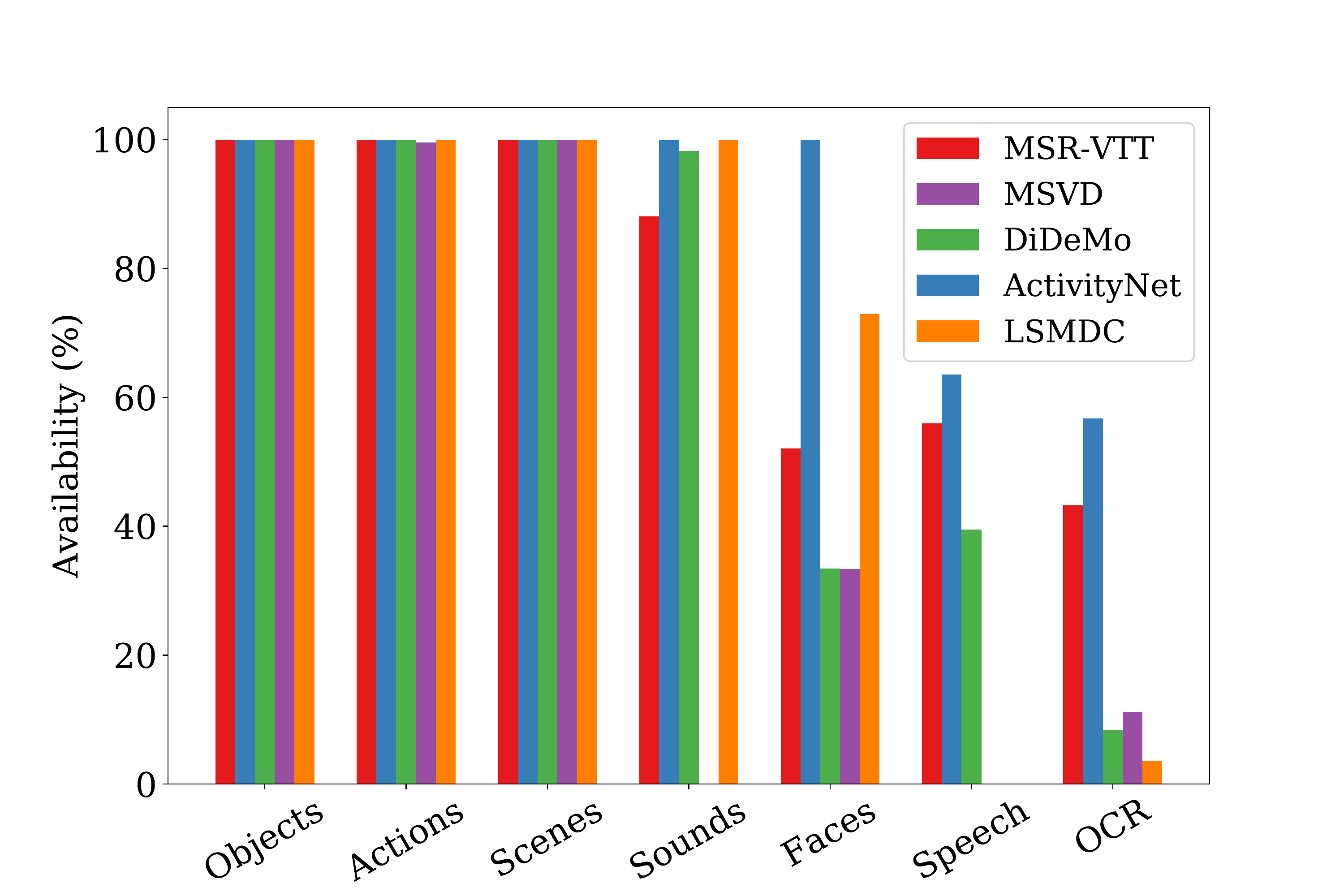}\\
        \end{subfigure} \\
    \end{minipage}
    \vspace{-0.8cm}
\caption{(Left): Unconstrained videos `in the wild' convey information in various different ways, including (clockwise from upper-left), clues from distinctive speech, names of individuals on screen, other text clues embedded in the video and audio. (Right): For the five video datasets considered in this work, the chart portrays the video-level availability of ``expert'' embeddings from different domains (with potentially multiple experts per domain): certain generic embeddings can almost always be extracted via pretrained object/action/scene classification networks. Other features such as sounds, faces, speech and OCR are less consistently available and are more challenging to exploit (Sec.~\ref{subsec:ablations}). \label{fig:teaser}}   
\vspace{-1em}
\end{figure}

Learning a robust and compact representation \textit{tabula rasa} for
this task is made extremely challenging by the high dimensionality of
the sensory data contained in videos---to do so with discriminative
training would require prohibitively expensive textual annotation of a
vast number of videos.  The primary hypothesis underpinning our
approach is the following: \textit{the discriminative content of the multi-modal video embedding
can be well approximated by the set of semantic representations of the video data learnt by individual experts (in audio, 
scenes, actions, etc)}. In essence, this approximation
enables us to exploit knowledge from
existing individual sources where the cost of annotation is significantly reduced
(e.g.\  classification labels for objects and scenes in images, labels for actions in
videos etc.) and where consequently, there exist very large-scale labelled datasets.
These large-scale datasets can then be
used to train independent \say{experts} for different 
perception tasks, which in turn provide a robust,
low-dimensional basis for the discriminative query-content
approximation described above. 

The two key aspects of this idea that we explore in this paper are: (i) {\em General and specific features:}  in addition to using generic video descriptors (e.g.\ objects and actions) we investigate encodings of quite specific information from the clip, for example, text from overlaid captions and text from speech to provide effective coverage of the ``queryable content'' of the video (Fig.~\ref{fig:teaser}, left). While such features may be highly discriminative for humans, they may not always be available (Fig.~\ref{fig:teaser}, right) and as we show through experiments (Sec.~\ref{subsec:ablations}), making good use of these cues is challenging. We therefore also propose (ii) {\em Collaborative experts:} a framework that seeks to make effective use of embeddings from different `experts' (e.g.\ objects, actions, speech) by learning their combination in order to render them more discriminative. Each expert is filtered via a simple dynamic attention mechanism that considers its relation to all other experts to enable their collaboration. This pairwise approach enables, for instance, the sound of a dog barking to inform the modulation of the RGB features, selecting the features that have encoded the concept of the dog. As we demonstrate in the sequel, this idea yields improvements in the retrieval performance.

Concretely, we make the following three contributions: 
(i) We propose the \textit{Collaborative Experts} framework for learning a joint embedding of video and text by combining a collection of pretrained embeddings into a single, compact video representation. Our joint video embeddings are independent of the retrieval text-query and can be pre-computed offline and indexed for efficient retrieval; 
(ii) We explore the use of both \textit{general} video features such as motion, image classification and audio features, and \textit{specific} video features such as text embedded on screen and speech obtained using OCR and ASR respectively. We find that strong generic features deliver good performance, but that specific, rarely available features remain challenging to use for retrieval\footnote{Note that this finding differs from the previous version of this paper (see appendix~\ref{sec:correction}).}.  
(iii) We assess the performance of the representation produced by combining all available cues on a number of retrieval benchmarks, in several cases achieving an advance over prior work.
% \vspace{-0.5cm}
\section{Related Work} 

\noindent\textbf{Cross-Modal Embeddings:} A range of prior work has proposed to jointly embed images and text into the same space~\cite{farhadi2010every,frome2013devise,faghri2017vse,kiros2014unifying,nam2017dual}, enabling cross-modal retrieval. More recently, several works have also focused on audio-visual cross-modal embeddings~\cite{arandjelovic2017look,nagrani2018learnable}, as well as audio-text embeddings~\cite{chechik2008large}. Our goal in this work, however, is to embed videos and natural language sentences (sometimes multiple sentences) into the same semantic space, which is made more challenging by the high dimensional content of videos.  \\ 
\noindent\textbf{Video-Text Embeddings:} 
While a large number of
works~\cite{dong2016word2visualvec,otani2016learning,pan2016jointly,torabi2016learning,xu2015jointly}
have focused on learning visual semantic embeddings for video and
language, many of these existing approaches are based on image-text
embedding methods by design and typically focus on single visual frames. Mithun el al.~\cite{mithun2018learning}
observe that a simple adaptation of a state-of-the-art image-text
embedding method~\cite{faghri2017vse} by mean-pooling features from
video frames provides a better result than many prior video-text
retrieval approaches~\cite{dong2016word2visualvec,otani2016learning}.
However, such methods do not take advantage of the rich and varied
additional information present in videos, including motion
dynamics, speech and other background sounds, which may influence the
concepts in human captions to a considerable extent.  Consequently, there has been a
growing interest in fusing information from other modalities---\cite{mithun2018learning,miech2018learning} utilise the audio stream
(but do not exploit speech content) and use models pretrained for
action recognition to extract motion features. These methods do not make use of speech-to-text or OCR for additional cues, which have nevertheless been used successfully to understand videos in other domains, particularly lecture retrieval~\cite{radha2016video,yamamoto2003topic} (where the videos consist of slide shows) and news broadcast~\cite{hauptmann2002multi} retrieval, where a large fraction of the content is displayed on screen in the form of text.   Our approach draws particular inspiration from the powerful joint embedding proposed by~\cite{miech2018learning} (which in turn, builds on the classical Mixtures-of-Experts model~\cite{jordan1994hierarchical}) and extends it to investigate additional cues (such as speech and text) and make more effective use of pretrained features via the robust collaborative gating mechanism described in Sec.~\ref{sec:collab-experts}.\\
\noindent\textbf{Annotation scarcity:} A key challenge for video-retrieval is the small size of existing training datasets, due to the high cost of annotating videos with natural language. We therefore propose to use the knowledge from existing embeddings pretrained on a wide variety of other tasks. This idea is not new: semantic projections of visual inputs in the form of `experts' was used by~\cite{douze2011combining} for the task of image retrieval and has also been central to modern video retrieval methods such as ~\citep{miech2018learning,mithun2018learning}.
More recently, alternative approaches to addressing the issue of annotation scarcity have been explored, which include self-supervised~\cite{sun2019videobert} and weakly-supervised~\cite{zhukov2019cross} video-text models. 
\section{Collaborative Experts} \label{sec:collab-experts}

\begin{figure}[!t]
\begin{center}
\includegraphics[width=1\textwidth,trim={0 0.25cm 0 0.25cm},clip]{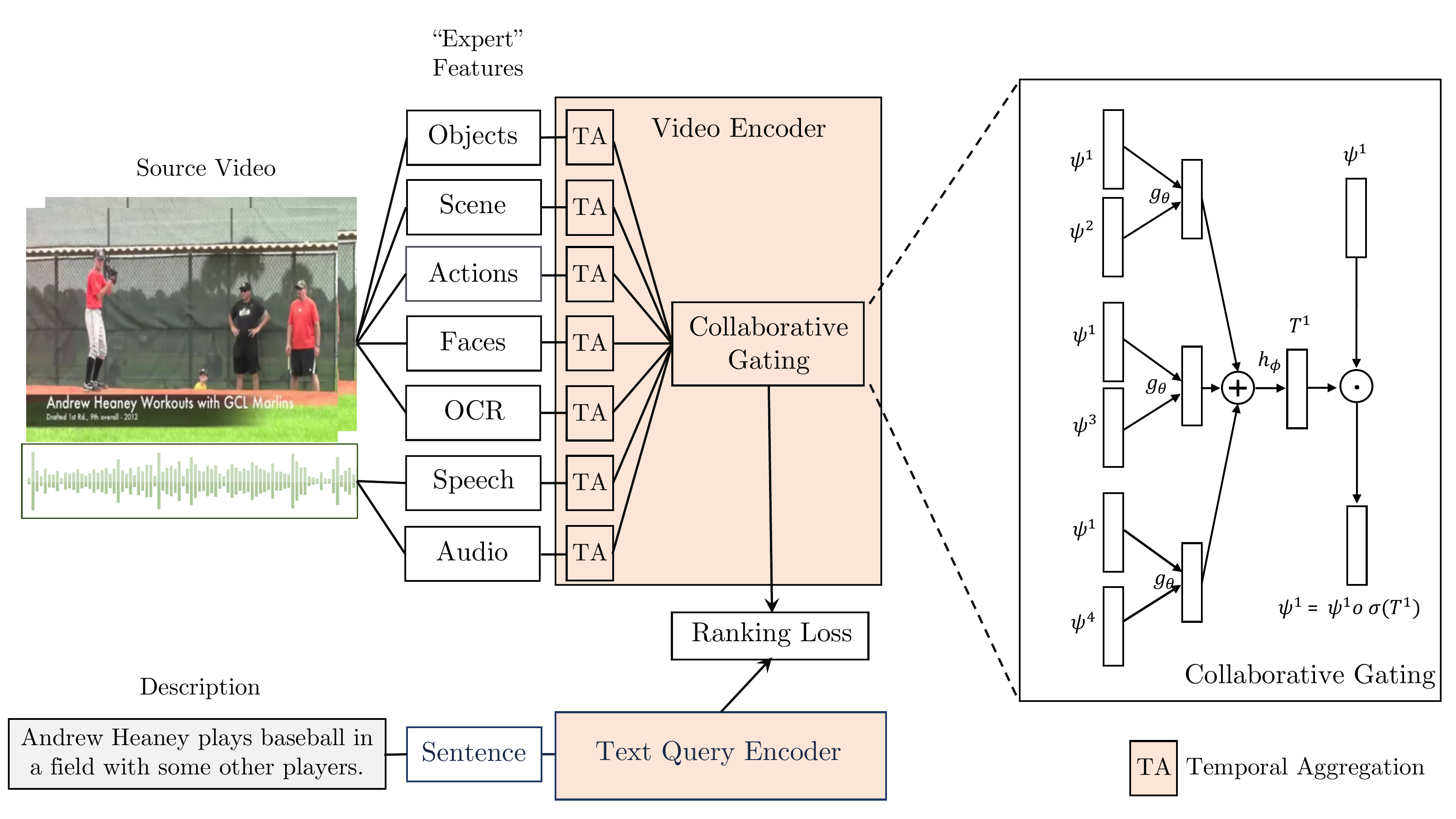}
\end{center}
  \caption{(Left): The proposed Collaborative Experts framework for learning a joint video-text embedding (coloured boxes denote learnable parameters). The information provided by each pretrained ``expert'' (potentially with multiple experts from a single domain) is temporally aggregated as it enters the video encoder and then refined through the use of a collaborative gating mechanism (right) to obtain the video-embedding (for visual clarity, we show the interaction of just a single expert with three others, though in practice all experts are used---see Sec.~\ref{subsec:video-encoder} for details). Note that to maintain retrieval efficiency, collaboration occurs only between video experts (the text-query and video embeddings are computed independently).\label{fig:CE-framework}}
\end{figure}

Given a set of videos with corresponding text captions, we would like to create a pair of functions $\phi_v$ and $\phi_t$  
that map sensory video data and text into a joint embedding space that respects this correspondence---embeddings for paired text and video should lie close together, while embeddings for text and video that do not match should lie far apart.  
We would also like $\phi_v$ and $\phi_t$ to be independent of each other to enable efficient retrieval: the process of querying then reduces to a distance comparison between the embedding of the query and the embeddings of the collection to be searched (which can be pre-computed offline). The proposed Collaborative Experts framework for learning these functions is illustrated in Fig.~\ref{fig:CE-framework}.  In this work, we pay particular attention to the design of the video encoder $\phi_v$ and the process of combining information from different video modalities (Sec.~\ref{subsec:video-encoder}). To complete the framework, we then discuss how the query text is encoded and the ranking loss used to learn the joint embedding space (Sec.~\ref{subsec:text-encoder}).

\subsection{Video Encoder} \label{subsec:video-encoder}
To construct the video encoder $\phi_v$, we draw on a collection of pretrained, single-modality experts.  These operate on the video sensory data $\bv$ and project it to a collection of $n$ variable-length task-specific embeddings $\{\Psi_{\text{var}}^{(1)}(\bv), \dots, \Psi_{\text{var}}^{(n)} (\bv)\}$. Here $\Psi_{\text{var}}^{(i)}$ represents the $i^{th}$ expert (we use the \say{var} subscript to denote a variable-length output when applied to a sequence of frames) whose parameters have been learned on a prior task such as object classification and then frozen.  Each element of this collection is then aggregated along its temporal dimension to produce fixed-size, task-specific embeddings per video $\{\Psi^{(1)}(\bv), \dots, \Psi^{(n)} (\bv)\}$.  Any temporal aggregation function may be used here---in this work, we use simple average pooling to aggregate \say{slow} visual features such as objects and scenes, and NetVLAD~\cite{arandjelovic2016netvlad} to aggregate more dynamic audio and word features (see Sec.~\ref{subsec:datsets} for further details). Next, to enable their combination, we apply linear projections to transform these task-specific embeddings to a common dimensionality. Our goal when fusing the resulting representations together into a single condensed video representation is to capture the valuable complementary information between task-specific projections while simultaneously filtering out irrelevant noise and resolving individual expert conflicts on a ~\textit{per-sample basis}. To do so, this we propose a collaborative gating module, described next. \\

\noindent\textbf{Collaborative Gating:} The collaborative gating module comprises two operations: (1) Prediction of attention vectors for every expert projection $T = \{T^{(1)}(\bv), \dots, T^{(n)} (\bv)\}$; and (2) modulation of expert responses.  Inspired by the relational reasoning module proposed by~\cite{santoro2017simple} for visual question answering, we define the attention vector of the $i^{th}$ expert projection $T_i$ as follows:
 \begin{equation} \label{eq_attention} 
 T^{(i)}(\bv) = h_\phi (\sum_{j\neq i} g_\theta (\Psi^{(i)}(\bv),\Psi^{(j)}(\bv))),
\end{equation}
where functions $h_\phi$ and $g_\theta$ are used to model the pairwise relationship between projection $\Psi^{(i)}$ and projection $\Psi^{(j)}$. Of these, $g_\theta$ is used to infer pairwise task relationships, while $h_\phi$ maps the sum of all pairwise relationships into a single attention vector.  In this work, we instantiate both $h_\phi$ and $g_\theta$ as multi-layer perceptrons (MLPs). Note that the functional form of Equation (\ref{eq_attention}) dictates that the attention vector of any expert projection should consider the potential relationships between all pairs associated with this expert. That is to say, the quality of each expert $\Psi^{(j)}$ should contribute in determining and selecting the information content from $\Psi^{(i)}$ in the final decision.  It is also worth noting that the collaborative gating module uses the same functions $g_\theta$ and $h_\phi$ (shared weights) to compute all pairwise relationships. This mode of operation encourages greater generalisation, since $g_\theta$ and $h_\phi$ are encouraged not to over-fit to features of any particular pair of tasks.  
After the attention vectors  $T = \{T^{(1)}(\bv), \dots, T^{(n)} (\bv)\}$ have been computed, each expert projection is modulated follows:

\begin{equation} 
 \Psi^{(i)}(\bv) =  \Psi^{(i)}(\bv) \circ \sigma ( T^{(i)}(\bv)),
\end{equation}

where $\sigma$ is an element-wise sigmoid activation and $\circ$ is the element-wise multiplication (Hadamard product). This gating function re-calibrates the strength of different activations of $ \Psi^{(i)}(\bv)$ and selects which information is highlighted or suppressed, providing the model with a powerful mechanism for dynamically filtering content from different experts.  A diagram of the mechanism is shown in Fig.~\ref{fig:CE-framework} (right).  The final video embedding is then obtained by passing the modulated responses of each expert through a Gated Embedding Module (GEM)~\cite{miech2018learning} (note that this operation produces l2-normalized outputs) before concatenating the outputs together into a single fixed-length vector.

\subsection{Text Query Encoder and Training Loss} \label{subsec:text-encoder}

 To construct the text embeddings, query sentences are first mapped to a sequence of feature vectors with pretrained contextual word-level embeddings (see Sec.~\ref{subsec:datsets} for details)---as with the video experts, the parameters of this first stage are frozen.  These are then aggregated, again using NetVLAD~\cite{arandjelovic2016netvlad}.  Following aggregation, we follow the text encoding architecture proposed by \cite{miech2018learning}, which projects the aggregated features to separate subspaces for each expert using GEMs (as with the video encoder, producing l2-normalized outputs).  Each projection is then scaled by a mixture weight (one scalar weight per expert projection), which is computed by applying a single linear layer to the aggregated text-features, and passing the result through a softmax to ensure that the mixture weights sum to one (see \cite{miech2018learning} for further details). Finally, the scaled outputs are concatenated, producing a vector of dimensionality that matches that of the video embedding. 

With the video encoder $\phi_v$ and text encoder $\phi_t$ as described, the similarity $s_i^j$ of the $i^{th}$ video, $\bv_i$, and the and $j^{th}$ caption, $\bt_j$, can then be directly computed as the cosine of the angle between their respective embeddings $\phi_v(\bv_i)^T \phi_t(\bt_j)$.  During optimisation, the parameters of the video encoder (including the collaborative gating module) and text query encoder (the coloured regions of Fig.~\ref{fig:CE-framework}) are learned jointly.  Training proceeds by sampling a sequence of minibatches of corresponding video-text pairs $\{\bv_i, \; \bt_i\}_{i=1}^{N_B}$ and minimising a \textit{Bidirectional Max-margin Ranking Loss} \cite{socher2014grounded}:

\begin{align}
    \mathcal{L}_{r} = \frac{1}{N_B} \sum_{i=1, j\neq i}^{N_B}  \text{max}(0, \; m + s_i^j - s_i^i) + \text{max}(0,\; m + s_j^i - s_i^i) \label{eqn:loss}
\end{align}
where $N_B$ is the batch size, and $m$ is a fixed constant which is set as a hyperparameter.  When assessing retrieval performance, at test time the embedding distances are simply computed via their inner product, as described above.

\subsubsection{Missing Experts} \label{subsec:missing-epxerts} When a set of expert features are missing, such as when there is no speech in the audio track, we simply zero-pad the missing experts when estimating the similarity score. To compensate for the implicit scaling introduced by missing experts (the similarity is effectively computed between shorter embeddings), we follow the elegant approach proposed by \cite{miech2018learning} and simply remove the mixture weights for missing experts, then renormalise the remaining weights such that they sum to one.  
 \section{Experiments}

In this section, we evaluate our model on five benchmarks for video retrieval tasks. The description of datasets, implementation details and evaluation metric are provided in Sec.~\ref{subsec:datsets}.  A comprehensive comparison on general video retrieval benchmarks is reported in Sec.~\ref{subsec:sota}.  We present an ablation study in Sec.~\ref{subsec:ablations} to explore how the performance of the proposed method is affected by different model configurations, including the aggregation methods, importance of different experts and number of captions in training.

\subsection{Datasets, Implementation Details and Metrics} \label{subsec:datsets}

\noindent\textbf{Datasets:} We perform experiments on five video datasets:  MSR-VTT~\cite{xu2016msr}, LSMDC~\cite{rohrbach2015dataset}, MSVD~\cite{chen2011collecting}, DiDeMo~\cite{anne2017localizing} and ActivityNet-captions~\cite{krishna2017dense}, covering a challenging set of domains which include videos from YouTube, personal collections and movies. \\
\noindent\textbf{Expert Features:} In order to capture the rich content of a video, we draw on existing powerful representations for a number of different semantic tasks. These are first extracted at a frame-level, then aggregated to produce a single feature vector per modality per video. 
\textit{RGB \say{object}} frame-level embeddings of the visual data are generated with two models: an SENet-154 model~\cite{hu2019squeeze} (pretrained on ImageNet for the task of image classification), and a ResNext-101~\cite{xie2017aggregated} pretrained on Instagram hashtags~\cite{mahajan2018exploring}. \textit{Motion} embeddings are generated using the I3D inception model~\cite{carreira2017quo} and a 34-layer R(2+1)D model~\cite{tran2018closer} trained on IG-65m~\cite{ghadiyaram2019large}. \textit{Face} embeddings are extracted in two stages: (1) Each frame is passed through an SSD face detector~\cite{liu2016ssd,opencv_library} to extract bounding boxes; (2) The image region of each box is passed through a ResNet50~\cite{he2016identity} that has been trained for the task of face classification on the VGGFace2 dataset~\cite{Cao18}. 
\textit{Audio} embeddings are obtained with a VGGish model, trained for audio classification on the YouTube-8m dataset \cite{hershey2017}. \textit{Speech-to-Text} features are extracted using the Google Cloud speech API, to extract word tokens from the audio stream, which are then encoded via pretrained word2vec embeddings~\cite{mikolov2013efficient}. \textit{Optical Character Recognition} is done in two stages: (1) Each frame is passed through the Pixel Link \cite{deng2018pixellink} text detection model to extract bounding boxes for text; (2) The image region of each box is passed through a model \cite{liu2018synthetically} that has been trained for scene text recognition on the Synth90K dataset\cite{jaderberg2014synthetic}. The text is then encoded via a pretrained word2vec embedding model ~\cite{mikolov2013efficient}. \\
\noindent \textbf{Temporal Aggregation:} We adopt a simple approach to aggregating the features described above.  For appearance, motion, scene and face embeddings, we average frame-level features along the temporal dimension to produce a single feature vector per video (we found max-pooling to perform similarly).  For speech, audio and OCR features, we adopt the NetVLAD mechanism proposed by \cite{arandjelovic2016netvlad}, which has proven effective in the retrieval setting \cite{miech2017learnable}. As noted in Sec.~\ref{subsec:video-encoder}, all aggregated features are projected to a common size (768 dimensions). \\
\textbf{Text:} Each word is encoded using pretrained word2vec word embeddings~\cite{mikolov2013efficient} and then passed through a pretrained OpenAI-GPT model~\cite{radford2018improving} to extract contextual word embeddings. Finally, the word embeddings in each sentence are aggregated using NetVLAD. \\
\noindent \textbf{Dataset-specific details:} Except where noted otherwise for ablation purposes, we use each of the embeddings described above for the MSR-VTT, ActivityNet and DiDeMo datasets. For MSVD, we extract the subset of features which do not require an audio stream (since no audio is available with the dataset).  For LSMDC, we re-use the existing face, text and audio features made available by \cite{miech2018learning}, and combine them with the remaining features described above. \\
\noindent \textbf{Training Details:} The CE framework is implemented with PyTorch~\cite{paszke2017automatic}.  Optimisation is performed with the Lookahead solver \cite{kingma2014adam} in combination with RAdam~\cite{liu2019variance} (implementation by~\cite{LessWright}).  Optimisation settings and the hyperparameter selection procedure is described in the appendix. \\
\noindent \textbf{Evaluation Metrics:} We follow prior work (e.g. ~\cite{dong2016word2visualvec,zhang2018cross,mithun2018learning,yu2018joint,miech2018learning}) and report standard retrieval metrics (where existing work enables comparison) including median rank (lower is better), mean rank (lower is better) and R@K (recall at rank K---higher is better).  When computing video-to-sentence metrics for datasets with multiple independent sentences per video (MSR-VTT and MSVD), we follow the evaluation protocol used in prior work~\cite{mithun2018learning,dong2018predicting,dong2018dual} which corresponds to reporting the minimum rank among all valid text descriptions for a given video query.  For each benchmark, we report the mean and standard deviation of three randomly seeded runs.

\begin{table}[t]
\centering 
\hspace{-0.2cm}
\footnotesize 
\setlength{\tabcolsep}{2pt}
\begin{tabular}{l | c |@{\hskip -0.25cm}c@{\hskip -0.35cm}c@{\hskip -0.35cm}c@{\hskip -0.1cm}c@{\hskip -0.2cm}c |@{\hskip -0.2cm}c@{\hskip -0.35cm}c@{\hskip -0.35cm}c@{\hskip -0.1cm}c@{\hskip -0.25cm}c} 
\hline \hline
\multicolumn{2}{c}{} & 
\multicolumn{5}{c}{Text $\implies$ Video} & \multicolumn{5}{c}{Video $\implies$ Text} \\
Method & Test-set & R$@$1 & R$@$5 & R$@$10 & MdR & MnR & R$@$1 & R$@$5 & R$@$10 & MdR & MnR \\ 
\hline 
JSFusion~\cite{yu2018joint} & 1k-A & 10.2 & 31.2 & 43.2 & 13 & - & - & - & - & - & - \\
\textbf{CE} & 1k-A & $\prepad \textbf{20.9}_{\pm1.2}$ & $\prepad \textbf{48.8}_{\pm0.6}$ & $\prepad \textbf{62.4}_{\pm0.8}$	&  $\prepadmini \textbf{6}_{\pm0}$ & $\prepad \textbf{28.2}_{\pm0.8}$ & $\prepad \textbf{20.6}_{\pm0.6}$	& $\prepad \textbf{50.3}_{\pm0.5}$	& $\prepad \textbf{64.0}_{\pm0.2}$& $\prepadmini \textbf{5.3}_{\pm0.6}$	& $\prepad \textbf{25.1}_{\pm0.8}$\\
\hline 
MoEE~\cite{miech2018learning}& 1k-B & 13.6  &  37.9  &  51.0  &  10 & - & - & - & - & - & - \\
MoEE$_{\text{COCO}}$~\cite{miech2018learning}& 1k-B & 14.2  &  39.2  &  53.8  &  9 & - & - & - & - & - & - \\
\textbf{CE} & 1k-B & $\prepad \textbf{18.2}_{\pm0.7}$	& $\prepad \textbf{46.0}_{\pm0.4}$	& $\prepad \textbf{60.7}_{\pm0.2}$ & $\prepadmini \textbf{7}_{\pm0}$	& $\prepad \textbf{35.3}_{\pm1.1}$   & $\prepad \textbf{18.0}_{\pm0.8}$	& $\prepad \textbf{46.0}_{\pm0.5}$	& $\prepad \textbf{60.3}_{\pm0.5}$& $\prepadmini \textbf{6.5}_{\pm0.5}$	& $\prepad \textbf{30.6}_{\pm1.2}$ \\
\hline 
VSE \cite{mithun2018learning} & Full & 5.0 & 16.4 & 24.6 & 47 & 215.1 & 7.7 & 20.3 & 31.2 & 28 & 185.8\\
VSE++ \cite{mithun2018learning} & Full & 5.7 & 17.1 & 24.8 & 65 & 300.8 & 10.2 & 25.4 & 35.1 & 25 & 228.1 \\
Mithun et al.~\cite{mithun2018learning} & Full &  7.0 & 20.9 & 29.7 & 38 & 213.8  & 12.5 & 32.1 & 42.4 & 16 & 134.0 \\
W2VV~\cite{dong2018predicting} & Full & 6.1 & 18.7 & 27.5 & 45 & - &  11.8 & 28.9 & 39.1 & 21 & -  \\  
Dual Enc.~\cite{dong2018dual} & Full & 7.7 & 22.0 & 31.8 & 32 & - &  13.0 & 30.8 & 43.3 & 15 & - \\
E2E ~\cite{miech2019end} & Full & 9.9 & 24.0 &  32.4 &  29.5 & - & - & - & - & -  & - \\
\textbf{CE} & Full & $\prepad \mathbf{10.0}_{\pm0.1}$	& $\prepad \mathbf{29.0}_{\pm0.3}$	& $\prepad \mathbf{41.2}_{\pm0.2}$	& $\prepadmini \mathbf{16}_{\pm0}$	& $\prepad \mathbf{86.8}_{\pm0.3}$ 	& $\prepad \mathbf{15.6}_{\pm0.3}$	& $\prepad \mathbf{40.9}_{\pm1.4}$	& $\prepad \mathbf{55.2}_{\pm1.0}$	& $\prepadmini \mathbf{8.3}_{\pm0.6}$ & $\prepad \mathbf{38.1}_{\pm1.8}$ \\
\hline \hline
\end{tabular}
\vspace{0.2cm}
\caption{Retrieval with sentences and videos on the MSR-VTT dataset. R$@$k denotes recall$@$k (higher is better), MdR and MnR denote median rank and mean rank resp. (lower is better). Standard deviations are reported from three randomly seeded runs. 1k-A and 1k-B denote test sets of 1000 randomly sampled text-video pairs used by~\cite{yu2018joint} and~\cite{miech2018learning} resp.}
\label{table:MSRVTT} 
\end{table}

\normalsize

\vspace{-0.2cm}
\subsection{Comparison to Prior State-of-the-Art} \label{subsec:sota}

We first compare the proposed method with the existing state-of-the-art on the MSR-VTT benchmark for the tasks of sentence-to-video and video-to-sentence retrieval Tab.~\ref{table:MSRVTT}.  Driven by strong expert features, we observe that Collaborative Experts (CE) consistently improves retrieval performance for both sentence and video queries.  We next evaluate the performance of the CE framework on the LSMDC benchmark for sentence-to-video retrieval (Tab.~\ref{table:LSMDC-MSVD}, left) and observe that CE matches or outperforms all prior work, including the prior state-of-the-art method \cite{miech2018learning} which incorporates additional training images and captions from the COCO benchmark during training, but uses fewer experts.  We observe similar trends in the results for the MSVD retrieval benchmark (Tab.~\ref{table:LSMDC-MSVD}, right). In Tab.~\ref{table:AN}, we compare with prior work on the ActivityNet paragraph-video retrieval benchmark (note that we compare to methods which use the same level of annotation as our approach i.e.\ video-level annotation), and see that CE is competitive.  Finally, in Tab.~\ref{table:DiDeMo} we provide a comparison with previously reported numbers on the DiDeMo benchmark and see that CE again outperforms prior work.

\begin{table}[ht]
\hspace{-0.6cm}
\begin{minipage}{0.52\linewidth}
\centering 
\scriptsize 
\setlength{\tabcolsep}{2pt}
\begin{tabular}{l | @{\hskip -0.2cm}c@{\hskip -0.35cm}c@{\hskip -0.35cm}c@{\hskip -0.1cm}c }
\hline \hline
\multicolumn{1}{c}{} & 
\multicolumn{4}{c}{Text $\implies$ Video} \\
Method & R$@$1 & R$@$5 & R$@$10 & MdR  \\ 
\hline 
Yu et al.~\cite{yu2016video}$\dagger$ &3.6 & 14.7 & 23.9 & 50 \\
CCA~\cite{klein2015associating} (rep. by~\cite{miech2018learning}) &7.5 & 21.7 & 31.0 & 33  \\
JSFusion~\cite{yu2018joint}$\ddagger$  & 9.1 & 21.2 & 34.1 & 36  \\
MoEE~\cite{miech2018learning} & 9.3 & 25.1 & 33.4 & 27 \\
$\text{MoEE}_\text{COCO}$~\cite{miech2018learning} & 10.1 & 25.6 & 34.6 & 27 \\
\textbf{CE} & $\prepad \textbf{11.2}_{\pm0.4}$ & $\prepad \textbf{26.9}_{\pm1.1}$	& $\prepad \textbf{34.8}_{\pm2.0}$ & $\prepad \textbf{25.3}_{\pm3.1}$	\\
\hline \hline
\end{tabular}
\end{minipage}
\begin{minipage}{0.4\linewidth}
\centering 
\scriptsize
\setlength{\tabcolsep}{4pt}
\begin{tabular}{l | @{\hskip -0.2cm}c@{\hskip -0.35cm}c@{\hskip -0.35cm}c@{\hskip -0.1cm}c@{\hskip -0.2cm}c } 
\hline \hline
\multicolumn{1}{c}{} & 
\multicolumn{5}{c}{Text $\implies$ Video}  \\
Method & R$@$1 & R$@$5 & R$@$10 & MdR & MnR  \\ 
\hline 
CCA (\cite{xu2015jointly}) & - & - & - & - & 245.3  \\
JMDV \cite{xu2015jointly} & - & - & - & - & 236.3  \\
VSE \cite{kiros2014unifying} (\cite{mithun2018learning}) & 12.3 & 30.1 & 42.3 & 14 & 57.7 \\
VSE++ \cite{faghri2017vse} (\cite{mithun2018learning}) & 15.4 & 39.6 & 53.0 & 9 & 43.8 \\
Multi. Cues \cite{mithun2018learning}  & $\mathbf{20.3}$ & 47.8 & 61.1 & $\mathbf{6}$ & 28.3 \\
\textbf{CE } & $\prepad 19.8_{\pm0.3}$	& $\prepad \mathbf{49.0}_{\pm0.3}$	& $\prepad \mathbf{63.8}_{\pm0.1}$	& $\prepad \mathbf{6}_{\pm0.0}$	& $\prepad \mathbf{23.1}_{\pm0.3}$ \\
\hline \hline
\end{tabular}
\end{minipage}
\vspace{0.2cm}
\caption{Text-to-Video retrieval results on the LSMDC dataset (left) and the MSVD dataset (right). $\dagger,\ddagger$ denote the winners of the 2016 and 2017 LSMDC challenges, respectively.}
\label{table:LSMDC-MSVD} 
\end{table}

\begin{table}[h!]
\centering 
\footnotesize 
\setlength{\tabcolsep}{2pt}
\begin{tabular}{l | @{\hskip -0.2cm}c@{\hskip -0.35cm}c@{\hskip -0.35cm}c@{\hskip -0.1cm}c@{\hskip -0.2cm}c |@{\hskip -0.2cm}c@{\hskip -0.35cm}c@{\hskip -0.35cm}c@{\hskip -0.1cm}c@{\hskip -0.2cm}c} 
\hline \hline
\multicolumn{1}{c}{} & 
\multicolumn{5}{c}{Text $\implies$ Video} & \multicolumn{5}{c}{Video $\implies$ Text} \\
Method & R$@$1 & R$@$5 & R$@$50 & MdR & MnR & R$@$1 & R$@$5 & R$@$50 & MdR & MnR \\ 
\hline 
S2VT~\cite{venugopalan2014translating} (\cite{zhang2018cross}) & 11.9 & 33.6 & 76.5 & 13 & - & 13.2 & 33.6 & 76.5 & 15 & - \\
FSE~\cite{zhang2018cross} & $\prepad 13.9_{\pm0.7}$ & $\prepad 36_{\pm0.8}$ & $\prepad {78.9}_{\pm 1.6}$ & $11$ & - & $\prepad {13.1}_{\pm0.5}$ & $\prepad {33.9}_{\pm0.4}$ & $\prepad 78.0_{\pm0.8}$ & $12$ &  - \\
\textbf{CE} & $\prepad \textbf{16.1}_{\pm1.4}$	& $\prepad \textbf{41.1}_{\pm0.4}$	& $\prepad \textbf{82.7}_{\pm0.3}$	& $\prepad \textbf{8.3}_{\pm0.6}$ & $\prepad \textbf{43.7}_{\pm3.6}$ & $\prepad \textbf{15.6}_{\pm1.3}$	& $\prepad \textbf{40.9}_{\pm0.4}$	& $\prepad \textbf{82.2}_{\pm1.3}$	& $\prepad \textbf{8.2}_{\pm0.3}$	& $\prepad \textbf{42.4}_{\pm3.3}$ \\
\hline \hline
\end{tabular}
\vspace{0.2cm}
\caption{Comparison of paragraph-video retrieval methods trained with video-level information on the DiDeMo dataset.}
\label{table:DiDeMo} 
\end{table}

\begin{table}[h]
% \captionsetup{font=footnotesize}
\centering 
\footnotesize 
\setlength{\tabcolsep}{2pt}
\begin{tabular}{l | @{\hskip -0.2cm}c@{\hskip -0.35cm}c@{\hskip -0.35cm}c@{\hskip -0.1cm}c@{\hskip -0.2cm}c |@{\hskip -0.2cm}c@{\hskip -0.35cm}c@{\hskip -0.35cm}c@{\hskip -0.1cm}c@{\hskip -0.2cm}c} 
\hline \hline 
\multicolumn{1}{c}{} & 
\multicolumn{5}{c}{Text $\implies$ Video} & \multicolumn{5}{c}{Video $\implies$ Text} \\
Method & R$@$1 & R$@$5 & R$@$50 & MdR & MnR & R$@$1 & R$@$5 & R$@$50 & MdR & MnR \\ 
\hline 
LSTM-YT~\cite{venugopalan2015sequence} (\cite{zhang2018cross}) & 0.0 & 4.0 & 24.0 & 102 & - & 0.0 & 7.0 & 38.0  & 98 & - \\
NOCTXT~\cite{venugopalan2014translating} (\cite{zhang2018cross})& 5.0 & 14.0 & 32.0 & 78 & - & 7.0 & 18.0 & 45.0 & 56 & - \\
DENSE~\cite{krishna2017dense} & 14.0 & 32.0 & 65.0 & 34 & - & 18.0 & 36.0 & 74.0 & 32 \\
FSE~\cite{zhang2018cross} & $\prepad 18.2_{\pm0.2}$ & $\prepad 44.8_{\pm0.4}$ & $\prepad {89.1}_{\pm 0.3}$ & $7$ & - & $\prepad {16.7}_{\pm0.8}$ & $\prepad {43.1}_{\pm1.1}$ & $\prepad 88.4_{\pm0.3}$ & $7$ &  - \\
HSE(4SEGS)~\cite{zhang2018cross}$\dagger$ & \textbf{20.5} & \textbf{49.3} & - & - & - & \textbf{18.7} & \textbf{48.1} & - & - \\
CE & $\prepad 18.2_{\pm0.3}$	& $\prepad 47.7_{\pm0.6}$	& $\prepad 91.4_{\pm0.4}$	& $\prepadmini 6_{\pm0}$	& $\prepad 23.1_{\pm0.5}$ & $\prepad 17.7_{\pm0.6}$	& $\prepad 46.6_{\pm0.7}$	& $\prepad 90.9_{\pm0.2}$	& $\prepadmini 6_{\pm0}$	& $\prepad 24.4_{\pm0.5}$ \\
\hline \hline
\end{tabular}
\vspace{0.3cm}
\caption{Comparison of paragraph-video retrieval methods trained with video-level information on the ActivityNet-captions dataset (val1 test-split).}
\label{table:AN} 
\end{table}

% \vspace{-0.5cm}
 \subsection{Ablation Studies} \label{subsec:ablations}

In this section, we provide ablation studies to empirically assess: (1) the effectiveness of the proposed collaborative experts framework vs other aggregation strategies; (2) the importance of using of a diverse range of experts with differing levels of specificity; (3) the relative value of using experts in comparison to simply having additional annotated training data.

\noindent \textbf{Aggregation method:} We compare the use of collaborative experts with several other baselines (with access to the same experts) for embedding aggregation including: (1) simple expert concatenation; (2) CE without projecting to a common dimension, without mixture weights and without the collaborative gating module described in Sec.~\ref{subsec:video-encoder}; (3) the state of the art MoEE~\cite{miech2018learning} method (equivalent to CE without the common projection and collaborative gating) and (4) CE without collaborative gating.  The results, presented in Tab.~\ref{table:agg-captions} (left), demonstrate the contribution of collaborative gating which improves performance and leads to a more efficient parameterisation than the prior state of the art. 

\noindent \textbf{Importance of different experts:} The value of different experts is assessed in Tab.~\ref{table:Ablation_single} (note that since several experts are not present in all videos, we combine them with features produced by a \say{scene} expert pretrained on Places365~\cite{zhou2017places}---the expert with the lowest performance that is consistently available as a baseline to enable a more meaningful comparison).  There is considerable variance in the effect produced by different choices of expert.  Using stronger features within a given modality (pretraining on Instagram~\cite{mahajan2018exploring}~rather than Kinetics~\cite{carreira2017quo} (resp.  ImageNet)~\cite{deng2009imagenet} for actions (resp. object) experts can yield a significant boost in performance).  The cues from scarce features (such as speech, face and OCR) which are often missing from videos (see Fig.~\ref{fig:teaser}, right) provide significantly weaker cues and bring a limited improvement to performance when used in combination. 
\begin{table}[h!]
\hspace{-0.25cm}
\begin{minipage}{0.45\linewidth}
\centering 
\scriptsize 
\setlength{\tabcolsep}{2pt}
\begin{tabular}{l c c c c c}
\hline \hline
\multicolumn{1}{c}{} & 
\multicolumn{5}{c}{Text $\implies$ Video}  \\
Experts & R$@$1 \hspace{0.5em} & R$@$5 \hspace{0.5em} & R$@$10\hspace{0.5em} & MdR\hspace{0.5em} & MnR\hspace{0.5em}  \\ 
\hline 

Scene &$ 4.0_{\pm0.1}$	&$ 14.1_{\pm0.1}$	&$ 22.4_{\pm0.3}$	&$ 50.0_{\pm1.0}$	&$ 201.3_{\pm1.6}$ \\
Scene+Speech &$ 4.6_{\pm0.1}$	&$ 15.5_{\pm0.2}$	&$ 24.4_{\pm0.2}$	&$ 44.7_{\pm1.2}$	&$ 183.6_{\pm1.7}$ \\
Scene+Audio &$ 5.6_{\pm0.0}$	&$ 18.7_{\pm0.1}$	&$ 28.2_{\pm0.1}$	&$ 33.7_{\pm0.6}$	&$ 140.8_{\pm0.3}$ \\
Scene+Action(KN) &$ 5.3_{\pm0.3}$	&$ 17.6_{\pm0.8}$	&$ 27.1_{\pm0.9}$	&$ 36.0_{\pm1.7}$	&$ 158.7_{\pm1.6}$ \\
Scene+Obj(IN) &$ 5.0_{\pm0.2}$	&$ 16.6_{\pm0.7}$	&$ 25.5_{\pm1.0}$	&$ 40.7_{\pm2.1}$	&$ 173.1_{\pm3.3}$ \\
Scene+Obj(IG) &$ 7.2_{\pm0.1}$	&$ 22.3_{\pm0.3}$	&$ 33.0_{\pm0.2}$	&$ 25.3_{\pm0.6}$	&$ 125.1_{\pm0.1}$ \\
Scene+Action(IG) &$ 6.8_{\pm0.1}$	&$ 21.7_{\pm0.1}$	&$ 32.4_{\pm0.1}$	&$ 25.7_{\pm0.6}$	&$ 122.1_{\pm0.3}$ \\
Scene+OCR	&$ 4.1_{\pm0.1}$	&$ 14.1_{\pm0.1}$	&$ 22.2_{\pm0.2}$	&$ 50.3_{\pm1.2}$	&$ 203.1_{\pm4.4}$ \\
Scene+Face &$ 4.1_{\pm0.1}$	&$ 14.2_{\pm0.3}$	&$ 22.4_{\pm0.4}$	&$ 49.7_{\pm0.6}$	&$ 194.2_{\pm5.1}$ \\
\hline \hline
\end{tabular}
\end{minipage}
\hspace{0.6cm}
\begin{minipage}{0.45\linewidth}
\centering 
\scriptsize 
\setlength{\tabcolsep}{2pt}
\begin{tabular}{l c c c c c}
\hline \hline
\multicolumn{1}{c}{} & 
\multicolumn{5}{c}{Text $\implies$ Video}  \\
Experts & R$@$1 \hspace{0.5em} & R$@$5 \hspace{0.5em} & R$@$10\hspace{0.5em} & MdR\hspace{0.5em} & MnR\hspace{0.5em}  \\ 
\hline 
Scene &$ 4.0_{\pm0.1}$	&$ 14.1_{\pm0.1}$	&$ 22.4_{\pm0.3}$	&$ 50.0_{\pm1.0}$	&$ 201.3_{\pm1.6}$ \\
Prev.+Speech &$ 4.6_{\pm0.1}$	&$ 15.5_{\pm0.2}$	&$ 24.4_{\pm0.2}$	&$ 44.7_{\pm1.2}$	&$ 183.6_{\pm1.7}$ \\
Prev.+Audio	&$ 5.8_{\pm0.1}$	&$ 19.0_{\pm0.3}$	&$ 28.8_{\pm0.2}$	&$ 32.3_{\pm0.6}$	&$ 136.8_{\pm1.2}$ \\
Prev.+Action(KN)	&$ 6.7_{\pm0.2}$	&$ 21.8_{\pm0.4}$	&$ 32.5_{\pm0.5}$	&$ 25.3_{\pm0.6}$	&$ 115.9_{\pm1.0}$ \\
Prev.+Obj(IN) &$ 7.5_{\pm0.1}$	&$ 23.4_{\pm0.0}$	&$ 34.1_{\pm0.2}$	&$ 23.7_{\pm0.6}$	&$ 111.9_{\pm0.6}$ \\
Prev.+Obj(IG)	&$ 9.5_{\pm0.2}$	&$ 27.7_{\pm0.1}$	&$ 39.4_{\pm0.1}$	&$ 18.0_{\pm0.0}$	&$ 92.6_{\pm0.4}$ \\
Prev.+Action(IG)	&$ 9.9_{\pm0.1}$	&$ 28.6_{\pm0.3}$	&$ 40.7_{\pm0.1}$	&$ 17.0_{\pm0.0}$	&$ 86.4_{\pm0.4}$ \\
Prev.+OCR	&$ 10.0_{\pm0.1}$	&$ 28.8_{\pm0.2}$	&$ 40.9_{\pm0.2}$	&$ 16.7_{\pm0.6}$	&$ 87.3_{\pm0.8}$ \\
Prev.+Face &$ 10.0_{\pm0.1}$	&$ 29.0_{\pm0.3}$	&$ 41.2_{\pm0.2}$	&$ 16.0_{\pm0.0}$	&$ 86.8_{\pm0.3}$ \\
\hline \hline

\end{tabular}
\end{minipage}
\vspace{0.2cm}
\caption{\textbf{The importance of different experts} (Left): The value of different experts in combination with a baseline set for text-video retrieval and (right) their cumulative effect on MSR-VTT (here Prev. denotes the experts used in the previous row).}
\label{table:Ablation_single} 
\end{table}

\begin{table}[ht]
\hspace{-0.25cm}
\begin{minipage}{0.45\linewidth}
\centering 
\scriptsize 
\setlength{\tabcolsep}{2pt}
\begin{tabular}{l c c c c c } 
\hline\hline
Aggreg. &  R$@$1 \hspace{0.5em} & R$@$5 \hspace{0.5em} & R$@$10\hspace{0.5em} & MdR\hspace{0.5em} & Params\hspace{0.5em}  \\ 
\hline 
Concat &$ 0.0_{\pm0.0}$	&$ 0.0_{\pm0.0}$	&$ 0.0_{\pm0.0}$	&$ 1495.5_{\pm0.0}$	& 369.72k \\
CE - MW,P,CG &$ 8.5_{\pm0.1}$	&$ 25.9_{\pm0.3}$	&$ 37.6_{\pm0.2}$	&$ 19.0_{\pm0.0}$ & 246.22M \\
MoEE~\cite{miech2018learning} &$ 9.6_{\pm0.1}$	&$ 28.0_{\pm0.2}$	&$ 39.7_{\pm0.2}$	&$ 17.7_{\pm0.6}$ & 400.41 M \\
CE - CG &$ 9.7_{\pm0.1}$	&$ 28.1_{\pm0.2}$	&$ 40.2_{\pm0.1}$	&$ 17.0_{\pm0.0}$ & 181.07 M \\
CE &$ 10.0_{\pm0.1}$	&$ 29.0_{\pm0.3}$	&$ 41.2_{\pm0.2}$	&$ 16.0_{\pm0.0}$ & 183.45 M \\
\hline\hline
\end{tabular}
\end{minipage}
\hspace{0.6cm}
\begin{minipage}{0.45\linewidth}

\centering 
\scriptsize 
\setlength{\tabcolsep}{2pt}
\begin{tabular}{l c| c  c c c}
\hline \hline
Expert & Num. Captions & R$@$1 & R$@$5 & R$@$10 & MdR   \\ 
\hline 
Obj(IN) & 1 &$ 2.6_{\pm0.1}$	&$ 9.3_{\pm0.4}$	&$ 15.0_{\pm0.7}$	&$ 101.3_{\pm15.5}$	\\
Obj(IN) & 20 &$ 4.9_{\pm0.1}$	&$ 16.5_{\pm0.2}$	&$ 25.3_{\pm0.4}$	&$ 40.7_{\pm1.2}$	\\
All & 1 &$ 4.8_{\pm0.2}$	&$ 16.2_{\pm0.5}$	&$ 25.0_{\pm0.7}$	&$ 43.3_{\pm4.0}$ \\
All & 20 &$ 10.0_{\pm0.1}$	&$ 29.0_{\pm0.3}$	&$ 41.2_{\pm0.2}$	&$ 16.0_{\pm0.0}$ \\
\hline \hline

\end{tabular}
\end{minipage}
\vspace{0.1cm}
\caption{(Left): Aggregation methods for text-video retrieval on MSR-VTT; (Right): The relative value of training with additional captions vs the value of experts.}
\label{table:agg-captions} 
\vspace{-3mm}
\end{table}

\begin{figure}[ht!]
\begin{center}
\vspace{-0.1cm}
\includegraphics[width=1.02\textwidth,trim={0cm 9.5cm 2.5cm 0.25cm},clip]{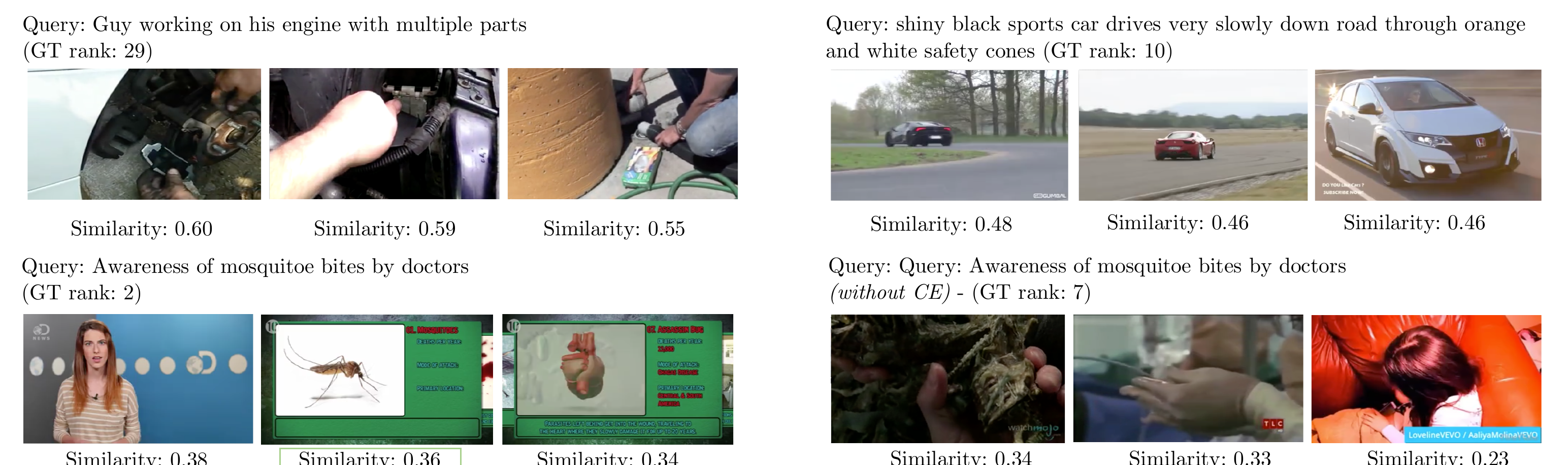}
\end{center}
\vspace{-1cm}
\hspace{-6mm}
  \caption{\textbf{Qualitative Results on MSR-VTT:} For each query, we show frames from the top three ranked videos (where present, the ground truth video is indicated by a green box around the similarity score). Top row: (left) Even for imperfect rankings, the model retrieves reasonable videos; Failure case (right) the embeddings can fail to differentiate between certain signals (in this case, ranking cars of the wrong colour above the ground truth video). Bottom row: (left) the videos retrieved by the proposed model (which assigns its second highest similarity to the correct video); (right) removing the proposed CE component produces a nosier ranking. \label{fig:qual-examples}}
  \vspace{-0.8cm}
\end{figure}

\noindent \textbf{Number of Captions in training:} An emerging idea in our community is that many machine perception tasks might be solved through the combination of simple models and large-scale training sets, reminiscent of the ``big-data'' hypothesis \cite{halevy2009unreasonable}. In this section, we perform an ablation study to assess the relative importance of access to pretrained experts and additional video description annotations.  To do so, we measure the performance of the CE model as we vary (1) the number of descriptions available per-video during training and (2) the number of experts it has access to. The results are shown in Tab.~\ref{table:agg-captions} (right). We observe that increasing the number of training captions per-video from 1 to 20 brings an improvement in performance, approximately comparable to adding the full collection of experts, suggesting that indeed, adding experts can help to compensate for a paucity of labelled data. When multiple captions and multiple experts are both available, they naturally lead to the most robust embedding.  Some qualitative examples of videos retrieved by the multiple-expert, multiple-caption system are provided in Fig.~\ref{fig:qual-examples}.

\section{Conclusion}

In this work, we introduced collaborative experts, a framework for learning a joint video-text embedding for efficient retrieval.  We have shown that using a range of pretrained features and combining them through an appropriate gating mechanism can boost retrieval performance. In future work, we plan to explore the use of collaborative experts for other video understanding tasks such as clustering and summarisation. 

\nopagebreak
% \vspace{-0.5cm}
 \noindent \paragraph{Acknowledgements:} Funding for this research is provided by the EPSRC Programme Grant Seebibyte EP/M013774/1 and EPSRC grant EP/R03298X/1. A.N. is supported by a Google PhD Fellowship. We would like to thank Antoine Miech, YoungJae Yu and Bowen Zhang for their assistance with experiment details.  We would like to particularly thank Valentin Gabeur for identifying a bug in the software implementation that was responsible for the inaccurate results reported in the initial version of the paper. We would also like to thank Zak Stone and Susie Lim for their help with cloud computing.

\clearpage
{\small
\bibliographystyle{ieee}
\bibliography{bibs/shortstrings,bibs/vgg_local,bibs/vgg_other,bibs/references}
}

\appendix
\section{Supplementary Material}

\subsection{Paper update, result corrections and summary of differences} \label{sec:correction}

Following the release of the initial version of this paper (which can be viewed for reference at \url{https://arxiv.org/abs/1907.13487v1}), a bug was discovered in our open-source software implementation which resulted in: (i) an overestimate of model performance; (ii) inaccurate conclusions about the relative importance of different experts on retrieval performance.  

This correction to the paper contains repeats of each of the experiments reported in the initial paper, with the following changes: (1) the removal of the bug which affected previous results; (2) a systematic approach to hyperparameter selection (discussed in more detail below); (3) the inclusion of additional \say{expert} pretrained features (described in Sec.~\ref{subsec:abl-details}) to assess the influence of feature strength within a modality.  In addition to results, the written analysis has also been updated to reflect the corresponding changes in results.  The authors would like to express their gratitude to Valentin Gabeur who identified the bug in the software implementation and enabled this correction.

\noindent \textbf{Bug details}: The bug caused information about feature availability in the ground truth target video to become available to the query encoder during both training and testing when computing embedding distances.  The leak occurred through incorrect weighting of the embedding distances due to: (1) a leaking broadcasting operation in an existing open-source library~\cite{miech2018learning} that was imported into our codebase; (2) incorrect NaN handling (introduced in our codebase), producing the same effect.  The bug has now been patched in each of the open-source codebases that were known to have used this implementation.

\subsection{Detailed Description of Datasets}
\textbf{MSR-VTT~\cite{xu2016msr}:} This large-scale dataset comprises approximately 200K unique video-caption pairs (10K YouTube video clips, each accompanied by 20 different captions).  The dataset is particularly useful because it contains a good degree of video diversity, but we noted a reasonably high degree of label noise (there are a number of duplicate annotations in the provided captions).  The dataset allocates 6513, 497 and 2990 videos for training, validation and testing, respectively. To enable a comparison with as many methods as possible, we also report results across other train/test splits used in prior work \cite{yu2018joint,miech2018learning}. In particular, when comparing with \cite{miech2018learning} (on splits which do not provide a validation set), we follow their evaluation protocol, measuring performance after training has occurred for a fixed number of epochs (100 in total).  \\
\textbf{MSVD \cite{chen2011collecting}:} The MSVD dataset %(also known as the YouTube2Text dataset) 
contains 80K English descriptions for 1,970 videos sourced from YouTube with a large number of captions per video (around 40 sentences each). We use the standard split of 1,200, 100, and 670 
videos for training, validation, and testing~\cite{venugopalan2015sequence,xu2015jointly}\footnote{Note: referred to by \cite{mithun2018learning} as the JMET-JMDV split}. Differently from the other datasets, the MSVD videos do not have audio streams. \\
\textbf{LSMDC~\cite{rohrbach2015dataset}:} This dataset contains 118,081 short video clips extracted from
202 movies. Each video has a caption, either extracted from the movie script or from transcribed 
DVS (descriptive video services) for the visually impaired.  The validation set contains 7408 clips and evaluation is performed on a test set of 1000 videos from movies disjoint from the training and val sets, as outlined by the Large Scale Movie Description
Challenge (LSMDC). \footnote{https://sites.google.com/site/describingmovies/lsmdc-2017} \\
\textbf{ActivityNet-captions~\cite{krishna2017dense}:}  ActivityNet Captions consists of 20K videos from YouTube, coupled with approximately 100K descriptive sentences. We follow the paragraph-video retrieval protocols described in \cite{zhang2018cross} training up to 200 epochs and reporting performance on  \texttt{val1} (this train/test split allocates 10,009 videos for training and 4,917 videos for testing). \\
\textbf{DiDeMo \cite{anne2017localizing}:} DiDeMo contains 10,464 unedited, personal videos in diverse visual settings with roughly 3-5 pairs of descriptions and distinct moments per video. The videos are collected in an open-world setting and include diverse content such as pets, concerts, and sports games. The total number of sentences is 40,543. While the moments are localised with time-stamp annotations, we do not use time stamps in this work. \\

\subsection{Optimisation details and hyperparameter selection}

For each dataset, a grid search was first performed (using the Lookahead solver \cite{zhang2019lookahead,LessWright}) over batch sizes (16, 32, 64, 128, 256), learning rates (0.1, 0.01) and weight decay (1E-3, 5E-5) for each dataset using a single expert to determine appropriate optimisation parameters.  Next, an experiment on MSR-VTT compared several choices for the dimensionality of the projection operation applied to the features (described in Sec.~\ref{subsec:video-encoder}) (choosing among 512, 768 and 1024 dimensions), which suggested that 768 was most effective.  This was then fixed for all remaining experiments (this represents a difference from the original paper, in which 512 was used). Further ablations (provided below) indicate that performance is not sensitive to this hyperparameter.  Next, Asynchronous Hyperband~\cite{li2018massively} was used to select all remaining hyperparameters on MSR-VTT by partially evaluating 1k configurations on the validation sets for each dataset.  These hyperparameters consisted of: the number of VLAD clusters and ghost clusters~\cite{zhong2018ghostvlad} used for different experts, the zero-padding length applied to variable-length experts, the margin hyperparameter $m$ in Eq.~\ref{eqn:loss}, the Collaborative Gating architecture (whether to use batch normalization~\cite{ioffe2015batch}, the number of layers used to form the MLP, and the choice of activation function).  The architecture choices were then fixed for all datasets.  Note that to ensure a fair comparison on MSR-VTT with the MoEE method of~\cite{miech2018learning} in Tab.~\ref{table:agg-captions}, MoEE was also provided with a budget of 1k sampled configurations. To determine zero-padding, margin and VLAD clusters for DiDeMo, MSVD and LSMDC further Asynchronous Hyperband searches were conducted, each with a budget of 500 sampled configurations.  Since, differently from the other datasets with available validation and test sets, the validation set itself is used to assess performance on ActivityNet, hyperparameters were copied from the DiDeMo configuration.  The configurations, experts, pretrained models and logs for each of the experiments reported in this paper are made available as part of the updated open-source implementation at \url{www.robots.ox.ac.uk/~vgg/research/collaborative-experts/}. 

\subsection{Ablation Studies - Full Tables}
\begin{table}[h!]
\centering 
\footnotesize 
\setlength{\tabcolsep}{2pt}
\begin{tabular}{l c c c c c| c c c c c}
\hline \hline
\multicolumn{1}{c}{} & 
\multicolumn{5}{c}{Text $\implies$ Video} & \multicolumn{5}{c}{Video $\implies$ Text} \\
Experts & R$@$1 & R$@$5 & R$@$10 & MdR & MnR & R$@$1 & R$@$5 & R$@$10 & MdR & MnR \\ 
\hline 
Scene &$ 4.0_{\pm0.1}$	&$ 14.1_{\pm0.1}$	&$ 22.4_{\pm0.3}$	&$ 50.0_{\pm1.0}$	&$ 201.3_{\pm1.6}$ &$ 5.6_{\pm0.6}$	&$ 18.2_{\pm0.6}$	&$ 27.7_{\pm0.3}$	&$ 39.0_{\pm0.0}$	&$ 247.0_{\pm10.1}$ \\
Scene+Speech &$ 4.6_{\pm0.1}$	&$ 15.5_{\pm0.2}$	&$ 24.4_{\pm0.2}$	&$ 44.7_{\pm1.2}$	&$ 183.6_{\pm1.7}$ &$ 6.0_{\pm0.2}$	&$ 20.4_{\pm0.5}$	&$ 30.3_{\pm1.0}$	&$ 33.0_{\pm2.0}$	&$ 222.6_{\pm9.9}$ \\
Scene+Audio &$ 5.6_{\pm0.0}$	&$ 18.7_{\pm0.1}$	&$ 28.2_{\pm0.1}$	&$ 33.7_{\pm0.6}$	&$ 140.8_{\pm0.3}$ &$ 8.2_{\pm0.4}$	&$ 24.8_{\pm0.4}$	&$ 36.0_{\pm0.1}$	&$ 21.7_{\pm0.6}$	&$ 127.9_{\pm5.9}$ \\
Scene+Action(KN) &$ 5.3_{\pm0.3}$	&$ 17.6_{\pm0.8}$	&$ 27.1_{\pm0.9}$	&$ 36.0_{\pm1.7}$	&$ 158.7_{\pm1.6}$ &$ 7.3_{\pm0.6}$	&$ 22.3_{\pm1.4}$	&$ 33.4_{\pm1.7}$	&$ 25.2_{\pm2.0}$	&$ 151.7_{\pm11.6}$ \\
Scene+Obj(IN) &$ 5.0_{\pm0.2}$	&$ 16.6_{\pm0.7}$	&$ 25.5_{\pm1.0}$	&$ 40.7_{\pm2.1}$	&$ 173.1_{\pm3.3}$ &$ 6.9_{\pm0.5}$	&$ 21.2_{\pm0.9}$	&$ 31.1_{\pm1.9}$	&$ 28.7_{\pm3.8}$	&$ 188.3_{\pm4.7}$ \\
Scene+Obj(IG) &$ 7.2_{\pm0.1}$	&$ 22.3_{\pm0.3}$	&$ 33.0_{\pm0.2}$	&$ 25.3_{\pm0.6}$	&$ 125.1_{\pm0.1}$ &$ 10.1_{\pm0.3}$	&$ 29.7_{\pm0.5}$	&$ 41.9_{\pm0.7}$	&$ 15.2_{\pm0.9}$	&$ 91.3_{\pm2.4}$ \\
Scene+Action(IG) &$ 6.8_{\pm0.1}$	&$ 21.7_{\pm0.1}$	&$ 32.4_{\pm0.1}$	&$ 25.7_{\pm0.6}$	&$ 122.1_{\pm0.3}$ &$ 9.4_{\pm0.3}$	&$ 27.8_{\pm0.6}$	&$ 40.1_{\pm1.1}$	&$ 17.2_{\pm1.1}$	&$ 87.8_{\pm4.2}$ \\
Scene+OCR	&$ 4.1_{\pm0.1}$	&$ 14.1_{\pm0.1}$	&$ 22.2_{\pm0.2}$	&$ 50.3_{\pm1.2}$	&$ 203.1_{\pm4.4}$ &$ 5.4_{\pm0.5}$	&$ 18.6_{\pm1.2}$	&$ 26.6_{\pm1.2}$	&$ 40.0_{\pm1.0}$	&$ 292.6_{\pm9.9}$ \\
Scene+Face &$ 4.1_{\pm0.1}$	&$ 14.2_{\pm0.3}$	&$ 22.4_{\pm0.4}$	&$ 49.7_{\pm0.6}$	&$ 194.2_{\pm5.1}$ &$ 5.6_{\pm1.0}$	&$ 17.9_{\pm0.7}$	&$ 26.7_{\pm0.8}$	&$ 39.1_{\pm2.6}$	&$ 273.5_{\pm6.3}$ \\

\hline \hline

\end{tabular}
\caption{Ablation study of importance of each expert when combined with Scene features.}
\label{table:Ablation_single_extended} 
\end{table}

\begin{table}[h!]
\centering 
\footnotesize 
\setlength{\tabcolsep}{2pt}
\begin{tabular}{l c c c c c| c c c c c}
\hline \hline
\multicolumn{1}{c}{} & 
\multicolumn{5}{c}{Text $\implies$ Video} & \multicolumn{5}{c}{Video $\implies$ Text} \\
Experts & R$@$1 & R$@$5 & R$@$10 & MdR & MnR & R$@$1 & R$@$5 & R$@$10 & MdR & MnR \\ 
\hline 
Scene &$ 4.0_{\pm0.1}$	&$ 14.1_{\pm0.1}$	&$ 22.4_{\pm0.3}$	&$ 50.0_{\pm1.0}$	&$ 201.3_{\pm1.6}$ &$ 5.6_{\pm0.6}$	&$ 18.2_{\pm0.6}$	&$ 27.7_{\pm0.3}$	&$ 39.0_{\pm0.0}$	&$ 247.0_{\pm10.1}$ \\
Prev.+Speech &$ 4.6_{\pm0.1}$	&$ 15.5_{\pm0.2}$	&$ 24.4_{\pm0.2}$	&$ 44.7_{\pm1.2}$	&$ 183.6_{\pm1.7}$ &$ 6.0_{\pm0.2}$	&$ 20.4_{\pm0.5}$	&$ 30.3_{\pm1.0}$	&$ 33.0_{\pm2.0}$	&$ 222.6_{\pm9.9}$ \\
Prev.+Audio &$ 5.8_{\pm0.1}$	&$ 19.0_{\pm0.3}$	&$ 28.8_{\pm0.2}$	&$ 32.3_{\pm0.6}$	&$ 136.8_{\pm1.2}$ &$ 8.6_{\pm0.2}$	&$ 26.1_{\pm0.6}$	&$ 37.8_{\pm0.8}$	&$ 19.8_{\pm0.8}$	&$ 117.7_{\pm2.9}$ \\
Prev.+Action(KN) &$ 6.7_{\pm0.2}$	&$ 21.8_{\pm0.4}$	&$ 32.5_{\pm0.5}$	&$ 25.3_{\pm0.6}$	&$ 115.9_{\pm1.0}$ &$ 9.9_{\pm0.4}$	&$ 28.6_{\pm0.7}$	&$ 41.7_{\pm0.8}$	&$ 15.7_{\pm0.6}$	&$ 77.9_{\pm5.2}$ \\
Prev.+Obj(IN) &$ 7.5_{\pm0.1}$	&$ 23.4_{\pm0.0}$	&$ 34.1_{\pm0.2}$	&$ 23.7_{\pm0.6}$	&$ 111.9_{\pm0.6}$ &$ 11.2_{\pm0.3}$	&$ 32.1_{\pm0.8}$	&$ 45.4_{\pm0.6}$	&$ 13.7_{\pm0.6}$	&$ 68.0_{\pm1.4}$ \\
Prev.+Obj(IG) &$ 9.5_{\pm0.2}$	&$ 27.7_{\pm0.1}$	&$ 39.4_{\pm0.1}$	&$ 18.0_{\pm0.0}$	&$ 92.6_{\pm0.4}$ &$ 14.7_{\pm0.6}$	&$ 38.9_{\pm0.8}$	&$ 53.1_{\pm1.0}$	&$ 9.3_{\pm0.6}$	&$ 45.6_{\pm2.1}$ \\
Prev.+Action(IG) &$ 9.9_{\pm0.1}$	&$ 28.6_{\pm0.3}$	&$ 40.7_{\pm0.1}$	&$ 17.0_{\pm0.0}$	&$ 86.4_{\pm0.4}$ &$ 15.5_{\pm0.6}$	&$ 40.1_{\pm1.2}$	&$ 54.4_{\pm1.3}$	&$ 8.7_{\pm0.6}$	&$ 39.4_{\pm0.9}$	\\
Prev.+ OCR &$ 10.0_{\pm0.1}$	&$ 28.8_{\pm0.2}$	&$ 40.9_{\pm0.2}$	&$ 16.7_{\pm0.6}$	&$ 87.3_{\pm0.8}$ &$ 15.2_{\pm0.1}$	&$ 41.1_{\pm0.6}$	&$ 54.6_{\pm0.7}$	&$ 8.5_{\pm0.5}$	&$ 38.5_{\pm0.6}$ \\
Prev.+ Face &$ 10.0_{\pm0.1}$	&$ 29.0_{\pm0.3}$	&$ 41.2_{\pm0.2}$	&$ 16.0_{\pm0.0}$	&$ 86.8_{\pm0.3}$ &$ 15.6_{\pm0.3}$	&$ 40.9_{\pm1.4}$	&$ 55.2_{\pm1.0}$	&$ 8.3_{\pm0.6}$	&$ 38.1_{\pm1.8}$ \\

\hline \hline

\end{tabular}
\caption{Ablation study of the importance experts on the MSR-VTT dataset.}
\end{table}

\begin{table}[h!]
\centering 
\footnotesize 
\setlength{\tabcolsep}{1.5pt}
\begin{tabular}{l c| c c c c c| c c c c c}
\hline \hline
\multicolumn{1}{c}{} & \multicolumn{1}{c}{} &
\multicolumn{5}{c}{Text $\implies$ Video} & \multicolumn{5}{c}{Video $\implies$ Text} \\
Expert & Num. Caps & R$@$1 & R$@$5 & R$@$10 & MdR & MnR & R$@$1 & R$@$5 & R$@$10 & MdR & MnR \\ 
\hline 
Obj(IN) & 1 &$ 2.6_{\pm0.1}$	&$ 9.3_{\pm0.4}$	&$ 15.0_{\pm0.7}$	&$ 101.3_{\pm15.5}$	&$ 321.1_{\pm35.1}$ &$ 3.7_{\pm0.3}$	&$ 13.5_{\pm0.6}$	&$ 20.8_{\pm0.4}$	&$ 60.0_{\pm2.0}$	&$ 304.9_{\pm15.8}$ \\
Obj(IN) & 20 &$ 4.9_{\pm0.1}$	&$ 16.5_{\pm0.2}$	&$ 25.3_{\pm0.4}$	&$ 40.7_{\pm1.2}$	&$ 169.1_{\pm1.4}$ &$ 6.9_{\pm0.6}$	&$ 21.0_{\pm0.3}$	&$ 31.3_{\pm0.3}$	&$ 30.0_{\pm1.7}$	&$ 201.6_{\pm9.5}$ \\
All & 1 &$ 4.8_{\pm0.2}$	&$ 16.2_{\pm0.5}$	&$ 25.0_{\pm0.7}$	&$ 43.3_{\pm4.0}$	&$ 183.1_{\pm19.6}$ &$ 8.4_{\pm0.5}$	&$ 25.6_{\pm0.7}$	&$ 37.1_{\pm0.2}$	&$ 20.3_{\pm0.6}$	&$ 87.2_{\pm6.7}$ \\
All & 20 &$ 10.0_{\pm0.1}$	&$ 29.0_{\pm0.3}$	&$ 41.2_{\pm0.2}$	&$ 16.0_{\pm0.0}$	&$ 86.8_{\pm0.3}$ &$ 15.6_{\pm0.3}$	&$ 40.9_{\pm1.4}$	&$ 55.2_{\pm1.0}$	&$ 8.3_{\pm0.6}$	&$ 38.1_{\pm1.8}$ \\

\hline \hline

\end{tabular}
\caption{Ablation study of the number of captions in training on MSR-VTT}
\end{table}

\begin{table}[h!]
\centering 
\footnotesize 
\setlength{\tabcolsep}{1.5pt}
\begin{tabular}{l c c c c c| c c c c c | c}
\hline \hline
\multicolumn{1}{c}{} &
\multicolumn{5}{c}{Text $\implies$ Video} & \multicolumn{5}{c}{Video $\implies$ Text} & \multicolumn{1}{c}{} \\
Dimension & R$@$1 & R$@$5 & R$@$10 & MdR & MnR & R$@$1 & R$@$5 & R$@$10 & MdR & MnR & Params. \\ 
\hline   
384	&$ 9.4_{\pm0.2}$	&$ 27.8_{\pm0.4}$	&$ 39.8_{\pm0.4}$	&$ 17.7_{\pm0.6}$	&$ 88.8_{\pm0.5}$ &$ 14.0_{\pm0.5}$	&$ 38.7_{\pm0.5}$	&$ 52.7_{\pm1.4}$	&$ 9.3_{\pm0.6}$	&$ 41.8_{\pm1.0}$ & 88.62M\\
512	&$ 9.8_{\pm0.3}$	&$ 28.6_{\pm0.4}$	&$ 40.6_{\pm0.4}$	&$ 17.0_{\pm0.0}$	&$ 88.0_{\pm0.7}$ &$ 14.8_{\pm0.4}$	&$ 40.4_{\pm0.6}$	&$ 53.9_{\pm0.4}$	&$ 8.8_{\pm0.3}$	&$ 38.8_{\pm1.5}$ & 119.51M\\	
640	&$ 10.1_{\pm0.1}$	&$ 28.8_{\pm0.1}$	&$ 40.9_{\pm0.2}$	&$ 16.7_{\pm0.6}$	&$ 87.6_{\pm0.2}$ &$ 15.6_{\pm0.6}$	&$ 41.3_{\pm0.7}$	&$ 55.0_{\pm0.5}$	&$ 8.3_{\pm0.6}$	&$ 37.3_{\pm1.8}$ & 151.12M\\
768	&$ 10.0_{\pm0.1}$	&$ 29.0_{\pm0.3}$	&$ 41.2_{\pm0.2}$	&$ 16.0_{\pm0.0}$	&$ 86.8_{\pm0.3}$ &$ 15.6_{\pm0.3}$	&$ 40.9_{\pm1.4}$	&$ 55.2_{\pm1.0}$	&$ 8.3_{\pm0.6}$	&$ 38.1_{\pm1.8}$ & 183.45M\\	
1024 &$ 9.9_{\pm0.1}$	&$ 28.6_{\pm0.3}$	&$ 40.7_{\pm0.4}$	&$ 17.0_{\pm0.0}$	&$ 87.6_{\pm1.1}$ &$ 14.7_{\pm0.4}$	&$ 40.7_{\pm0.8}$	&$ 54.4_{\pm0.3}$	&$ 8.5_{\pm0.5}$	&$ 39.1_{\pm1.7}$ & 250.27M\\	
\hline \hline

\end{tabular}
\caption{Ablation study of the importance of model capacity by varying the shared embedding dimension used by CE on MSR-VTT.}
\end{table}

\subsection{Implementation Details} \label{subsec:abl-details}
\textbf{Object} frame-level embeddings of the visual data are generated with two models, \textit{Obj(IN)} and \textit{Obj(IG)}.  \textit{Obj(IN)} is an SENet-154 model \cite{hu2019squeeze} (pretrained on ImageNet for the task of image classification) from frames extracted at 25 fps, where each frame is resized to 224 $\times$ 224 pixels. \textit{Obj(IG)} is a ResNext-101~\cite{xie2017aggregated} pretrained on Instagram hashtags~\cite{mahajan2018exploring}, using the same frame preparation as \textit{Obj(IN)}.  Features are collected from the final global average pooling layer of both models, and have a dimensionality of 2048. \\ 
\textbf{Action} embeddings are similarly generated from two models, \textit{Action(KN)} and \textit{Action(IG)}.  \textit{Action(KN)} is an I3D inception model that computes features following the procedure described by \cite{carreira2017quo}.  Frames extracted at 25fps and processed with a window length of 64 frames and a stride of 25 frames.  Each frame is first resized to a height of 256 pixels (preserving aspect ratio), before a 224 $\times$ 224 centre crop is passed to the model.  Each temporal window produces a (1024x7)-matrix of features. \textit{Action(IG)} is a 34-layer R(2+1)D model~\cite{tran2018closer} trained on IG-65m~\cite{ghadiyaram2019large} which processes clips of 8 consecutive 112 $\times$ 112 pixel frames, extracted at 30 fps (we use the implementation provided by~\cite{Daniel}). \\ 
\textbf{Face} embeddings are extracted in two stages: (1) Each frame (also extracted at 25 fps) is resized to 300 $\times$ 300 pixels and passed through an SSD face detector~\cite{liu2016ssd,opencv_library} to extract bounding boxes; (2) The image region of each box is resized such that the minimum dimension is 224 pixels and a centre crop is passed through a ResNet50~\cite{he2016identity} that has been trained for task of face classification on the VGGFace2 dataset~\cite{Cao18}, producing a 512-dimensional embedding for each detected face. \\
\textbf{Audio} embeddings are obtained with a VGGish model, trained for audio classification on the YouTube-8m dataset \cite{hershey2017}. To produce the input for this model, the audio stream of each video is re-sampled to a 16kHz mono signal, converted to an STFT with a window size of 25ms and a hop of 10ms with a Hann window, then mapped to a 64 bin log mel-spectrogram.  Finally, the features are parsed into non-overlapping 0.96s collections of frames (each collection comprises 96 frames, each of 10ms duration), which is mapped to a 128-dimensional feature vector. \\
\textbf{Scene} embeddings of 2208 dimensions are extracted from 224$\times$224 pixel centre crops of frames extracted at 1fps using a DenseNet-161~\cite{huang2017densely} model pretrained on Places365~\cite{zhou2017places}.
\textbf{Speech to Text} The audio stream of each video is re-sampled to a 16kHz mono signal. We then obtained transcripts of the spoken speech for MSR-VTT, MSVD and ActivityNet using the Google Cloud Speech to Text API \footnote{https://cloud.google.com/speech-to-text/} from the resampled signal. The language for the API is specified as English. For reference, of the 10,000 videos contained in MSR-VTT, 8,811 are accompanied by audio streams.  Of these, we detected speech in 5,626 videos. \\
\textbf{Optical Character Recognition} is extracted in two stages: (1) Each frame is resized to $800 \times 400$ pixels) and passed through Pixel Link \cite{deng2018pixellink} text detection model to extract bounding boxes for texts; (2) The image region of each box is resized to $32 \times 256$ and then pass through a model \cite{liu2018synthetically,shi2017end} that has been trained for text of scene text recognition on the Synth90K dataset\cite{jaderberg2014synthetic}, producing a character sequence for each detect box. They are then encoded via a pretrained word2vec embedding model ~\cite{mikolov2013efficient}.\\
\textbf{Text} We encode each word using the Google News\footnote{
GoogleNews-vectors-negative300.bin.gz found at: https://code.google.com/archive/p/word2vec/} trained word2vec word embeddings~\cite{mikolov2013efficient}. All the word embeddings are then pass through a pretrained OpenAI-GPT model to extract the context-specific word embeddings (i.e., not only learned based on word concurrency but also the sequential context). Finally, all the word embeddings in each sentence are aggregated using NetVLAD. \\
 % Only used for arxiv version
\end{document}